\documentclass[a4paper, 10pt, conference]{ieeeconf}      

\IEEEoverridecommandlockouts                              

\usepackage{graphics} 
\usepackage{epsfig} 
\usepackage{mathptmx} 
\usepackage{times} 
\usepackage{amsmath} 
\usepackage{amssymb}  

\usepackage{color}
\usepackage{xcolor}
\usepackage{graphicx}
\usepackage{units}
\usepackage{epsfig}
\usepackage{hyperref}
\usepackage{subcaption}

\title{\LARGE \bf
	FLIC: Fast Lidar Image Clustering
}

\author{Frederik Hasecke$^{1,2}$, Lukas Hahn$^{1,2}$ and Anton Kummert$^{1}$
	\thanks{$^{1}$ University of Wuppertal, Germany - Faculty of Electrical Engineering\newline
		{ \tt\small firstname.lastname@uni-wuppertal.de}}
	\thanks{$^{2}$ Aptiv, Wuppertal, Germany\newline
		{ \tt\small firstname.lastname@aptiv.com}\vspace{0.4em}}
	\thanks{\footnotesize This work is a result of the research project @CITY – Automated Cars and Intelligent Traffic in the City. The project is supported by the Federal Ministry for Economic Affairs and Energy (BMWi), based on a decision taken by the German Bundestag. The author is solely responsible for the content of this publication.}}

\begin{document}

\maketitle
\thispagestyle{empty}
\pagestyle{empty}

\begin{abstract}
	
	Lidar sensors are widely used in various applications, ranging from scientific fields over industrial use to integration in consumer products. With an ever growing number of different driver assistance systems, they have been introduced to automotive series production in recent years and are considered an important building block for the practical realisation of autonomous driving. However, due to the potentially large amount of Lidar points per scan, tailored algorithms are required to identify objects (e.g. pedestrians or vehicles) with high precision in a very short time.\\
	In this work, we propose an algorithmic approach for real-time instance segmentation of Lidar sensor data. We show how our method leverages the properties of the Euclidean distance to retain three-dimensional measurement information, while being narrowed down to a two-dimensional representation for fast computation. We further introduce what we call \textit{skip connections}, to make our approach robust against over-segmentation and improve assignment in cases of partial occlusion. Through detailed evaluation on public data and comparison with established methods, we show how these aspects enable state-of-the-art performance and runtime on a single CPU core.
	
\end{abstract}

\section{Introduction}\label{sec:introduction}

\noindent Lidar sensors have been introduced in an ever growing number of different driver assistance systems to automotive series production in recent years and are considered an important building block for the practical realisation of autonomous driving. Precise segmentation of object instances is an important information for a variety of applications ranging from online object detection to offline computer-aided data labelling. Lidar sensor data is usually represented as a three-dimensional point cloud in Cartesian coordinates. Hence it makes sense to consider clustering algorithms to fulfil the task of object segmentation for this type of sensor. This work presents a Lidar point cloud segmentation approach, which provides a high level of accuracy in point cloud segmentation, while being able to run in real time, faster than the sensor recording frequencies at a constant rate with very little fluctuation independent of the scene's context. We do so by avoiding the creation of a three-dimensional point cloud from the range measurements provided by the Lidar scanner and work directly on the laser range values of the sensor.
\begin{figure}
	\centering
	\includegraphics[width=0.5\textwidth]{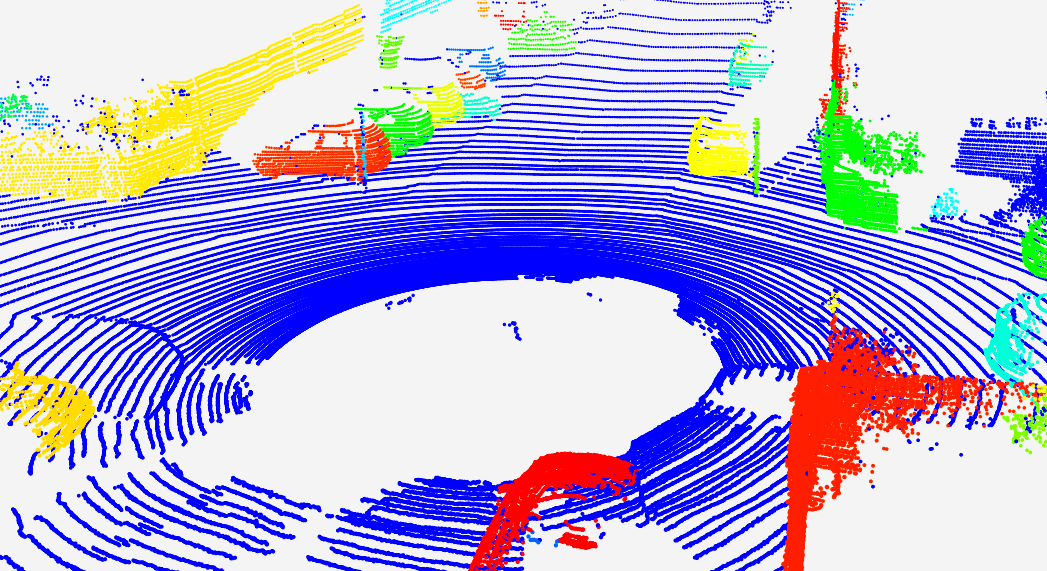}
	\caption{Three-dimensional point cloud with clustered points. Every Instance is assigned a random colour.}
	\label{fig:result}
\end{figure}
If not available, the computations can be applied to a spherical range image projection of the three-dimensional point cloud and projected back to the point cloud as demonstrated in Figure \ref{fig:result}. This approach circumvents the problem of sparsity in the point cloud by forcing a two-dimensional neighbourhood on each measurement and thus offers the advantage of working with dense, two-dimensional data with clearly defined neighbourhood relationships between adjacent measurements. A Python implementation of our approach runs in real time on a single Intel\textsuperscript{\textregistered} Core\texttrademark \: i7-6820HQ CPU @ 2.70 GHz core at a frame rate of up to 165 Hz.

\section{Related Work}\label{chapter:relatedwork}

\noindent There is a large number of previous works on Lidar instance segmentation especially of, but not limited to, automotive applications. The main focus of most clustering approaches is on improving segmentation accuracy and execution time. Most of these separate objects in the three-dimensional space, result in high accuracy but comparatively long runtime. Notable examples are the DBSCAN \cite{ester1996density}, Mean Shift \cite{fukunaga1975estimation} \cite{comaniciu2002mean} and OPTICS \cite{ankerst1999optics} algorithm.

Other approaches use voxelization to reduce the complexity of the point cloud and find clusters in the representation \cite{himmelsbach2010fast} or apply bird's eye view projection coupled with the height information to separate overlapping objects \cite{korchev2013real}.

\cite{moosmann2009segmentation} proposed the use of a local convexity criterion on the spanned surface of three-dimensional Lidar points in a graph-based approach. Based on this metric, \cite{Bogoslavskyi2016} used a similar criterion - the spanned angle between adjacent Lidar measurements in relation to the Lidar sensor origin - to define the convexity of the range image surface as a measure to segment objects. They further utilise the neighbourhood conditions in the range image to achieve the fastest execution time to date. \cite{zermas2017fast} exploit the same relationship in scan lines of Lidar sensors to find break points in each line and merge the separate lines of channels into three-dimensional objects in a subsequent step.

Other current methods use machine learning directly on three-dimensional point clouds \cite{lahoud20193d} \cite{zhang2020instance} \cite{yang2019learning}, projections into a camera image \cite{wang2019ldls}, or on spherical projections of Lidar points in the range image space \cite{wang2019pointit}, to segment object instances in point clouds. These results look very promising in some cases, but suffer from a longer runtime, which currently prevents application on embedded automotive hardware.

Related research areas that need to be mentioned at this point are the semantic segmentation and even more so the panoptic segmentation of Lidar data. The dataset ``SemanticKITTI"\cite{behley2019semantickitti}, which we use to evaluate our method in \ref{chapter:evaluation:quality}, has advanced a number of new algorithms in the area of semantic segmentation \cite{kochanov2020kprnet} \cite{zhou2020cylinder3d} \cite{zhang12356deep}. The task is concerned with the classification of each Lidar point. Leading algorithms have achieved great performance on assigning a class label and some are even real-time capable on a single GPU \cite{tang2020searching} \cite{gerdzhev2020tornado} \cite{cortinhal2020salsanext}. The difference to instance segmentation is the separation of instances. The semantic segmentation does not differentiate between multiple objects, but assigns the same label to each instance of the same class. The panoptic segmentation is the combination of the semantic and the instance segmentation. \cite{milioto2020iros} even manage to reach real-time capability on a single GPU with a runtime of \unit[85]{ms} by jointly segmenting the instances and the class.

Our method is concerned with the separation of instances independently of the class labels. We base our approach on the works of Bogoslavskyi and Stachniss \cite{Bogoslavskyi2016} as well as general three-dimensional Euclidean algorithms \cite{ester1996density} \cite{fukunaga1975estimation} \cite{comaniciu2002mean} \cite{ankerst1999optics}. We combine the fast execution time of range image clustering with the precise segmentation of distance thresholds to connect and separate Lidar points.

\section{Methods}\label{chapter:methods}

\noindent The raw data from Lidar sensors is usually provided as a list of range measurements, each coupled with a number relating to the channel and the lateral position. The two values correspond to the $y$- and $x$-position of the measurements in the range image. This can also be used to create a three-dimensional representation of the measurements, but doing so will increase the computation time for the proposed method. Therefore, we work directly on the raw two-dimensional representation.

\subsection{Ground Extraction}\label{chapter:methods:ground}

\begin{figure*}[tb]
	\centering
	\includegraphics[width=0.85\textwidth]{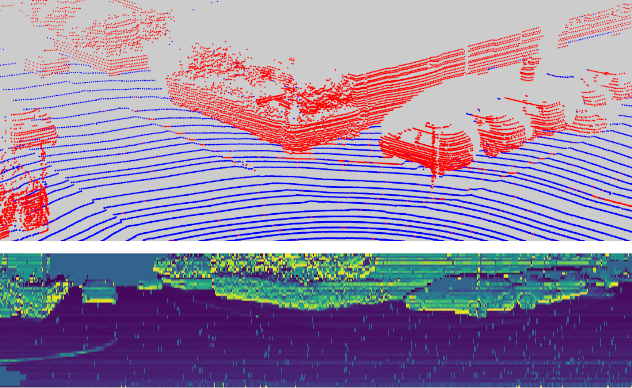}
	\caption{Top: Ground segmentation; Blue points represent ground points, red points are not part of the ground. \\Bottom: Angle image used to find horizontal surfaces for the ground segmentation.}
	\label{fig:ground_Extn}
\end{figure*}
For object segmentation, we assume that the Lidar is a part of a ground-based vehicle. Based on this assumption, we want to extract and ignore the points belonging to the ground plane from the segmentation. This will prevent the algorithm from connecting two instances via this plane. Here, a simple height based threshold is not sufficient, as the road surface itself can be uneven. Pitching and rolling of the ego vehicle can also influence the way the ground is perceived in the sensor data.
Since we have detailed information about the Lidar sensor itself, we can use the angle position of each given channel to determine the angle in which the laser beam would hit the surface. We use this information to exclude all range image values belonging to a horizontal plane below a certain height. To do so, we compare each cell of the range image with the neighbouring one above using the equation
\begin{equation}
\beta = \arctan{(\frac{d_2 \cdot \sin{\alpha}}{d_1 - d_2 \cdot \cos{\alpha}})},
\label{equation1}
\end{equation}
in which the selected cell value is $d_2$, the one above the cell is $d_1$ corresponding to the respective depth measurement and $\alpha$ is the angle between the two measurements.

\begin{figure}[tb]
	\centering
	\begin{subfigure}[t]{0.4\textwidth}
		\centering
		\includegraphics[width=\textwidth]{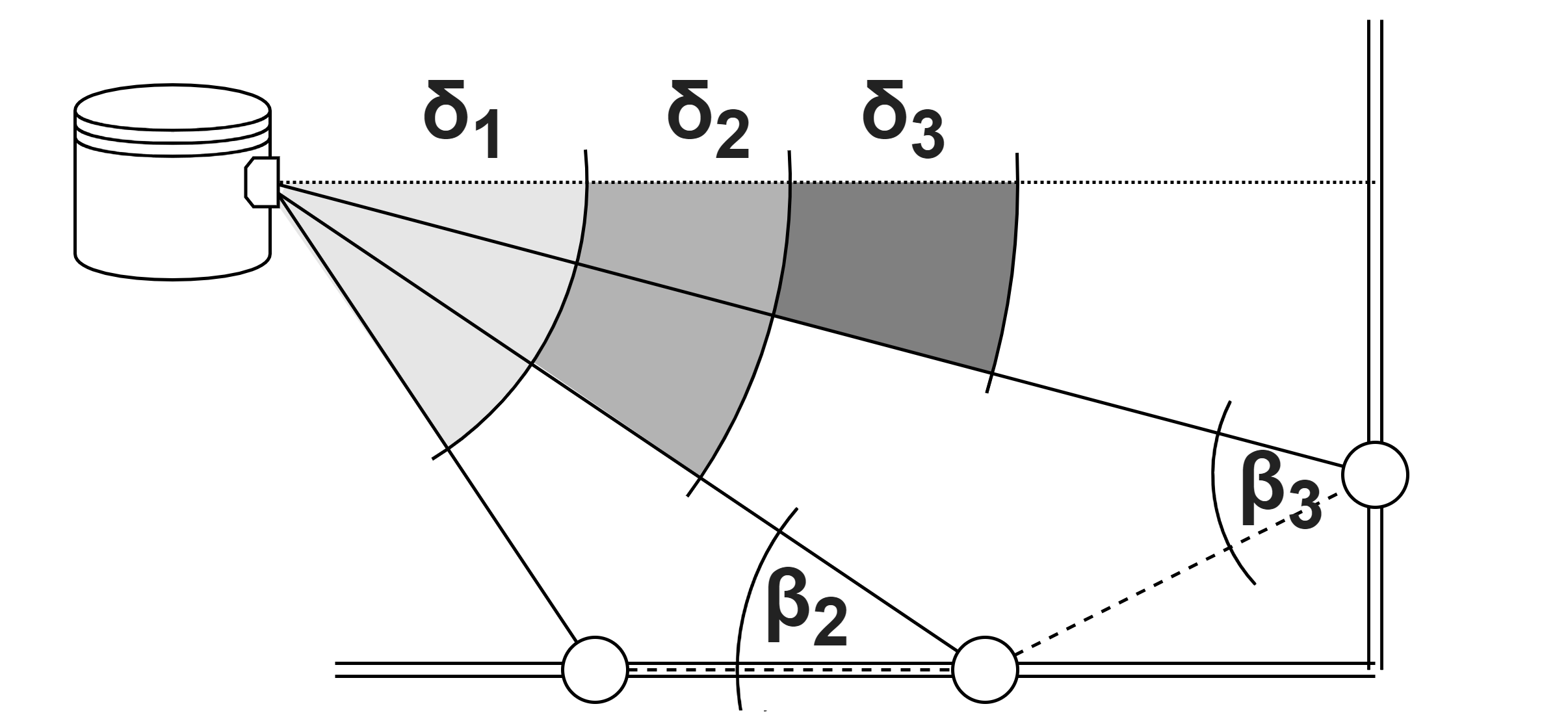}
		\caption{The angle measurement of Lidar points in the vertical direction is used to define a horizontal orientation for the ground plane extraction. (Note that there is not $\beta_1$, as the first Lidar channel has no previous channel, we therefore extend $\beta_2$ to the first channel.)}
	\end{subfigure}\hfill
	\begin{subfigure}[t]{0.4\textwidth}
		\centering
		\includegraphics[width=\textwidth]{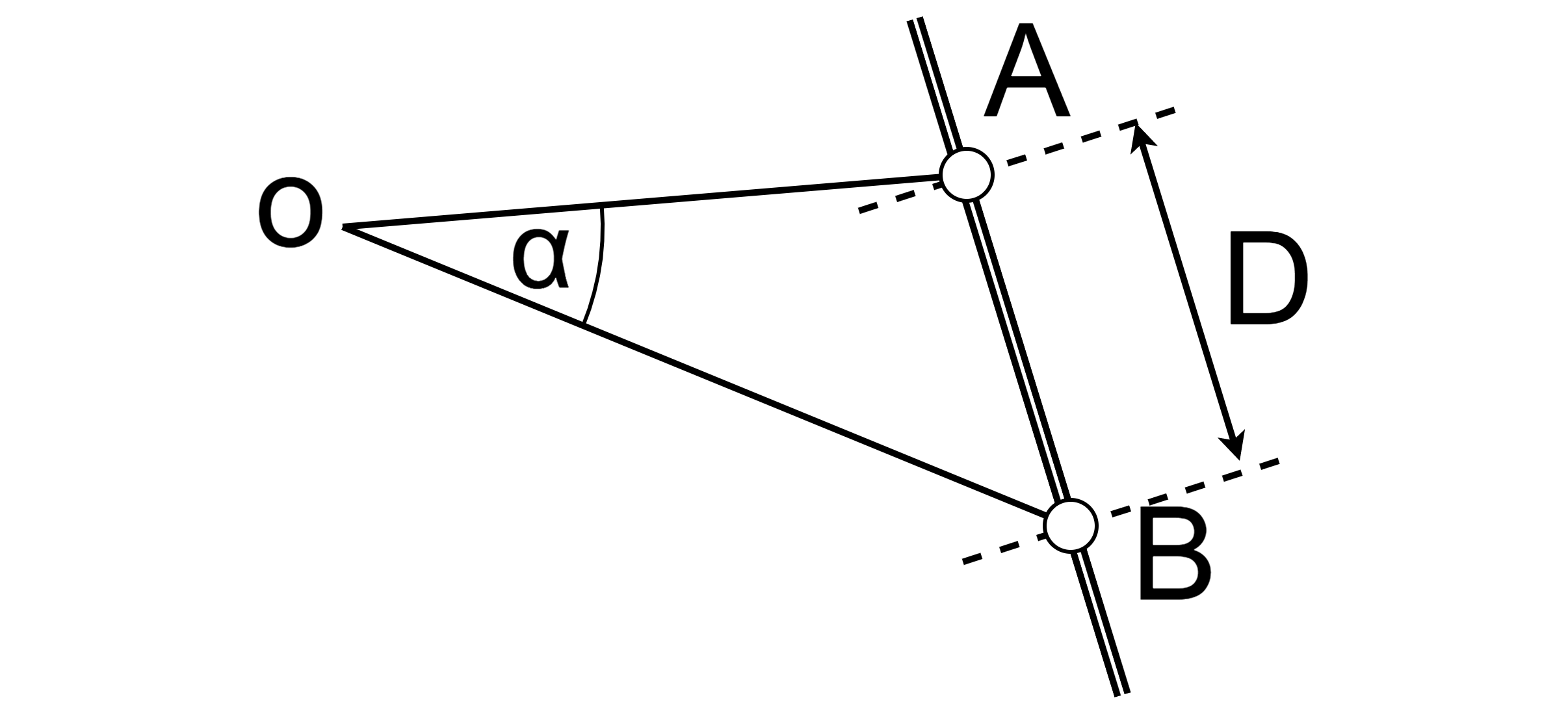}
		\caption{With the Lidar Sensor in $O$, the lines $OA$ and $OB$ show two neighbouring distance measurements. The distance between the two measurements is calculated using the spanned angle $\alpha$ between the points.\\}
	\end{subfigure}
	\caption{Trigonometric relationships used in the ground segmentation (a) and the cluster separation (b).\\}
	\label{groundAngle}
\end{figure}

As a result, we obtain the image shown in Figure \ref{fig:ground_Extn} representing the angle values of the Lidar beam in relation to the point cloud surface spanned between the current Lidar measurement and the measurement of the Lidar channel below, (see Figure \ref{groundAngle}). Using a lookup table of the channel angles $\delta_r$ with respect to the channel $r$, we exclude all range image cells that span a horizontal surface up to a certain threshold angle $\theta$ of $\pm 10^\circ$ to a full horizontal surface.

To speed up the computation we change the $\theta$ threshold to the tangent of $\theta$ to directly compare the tangent of $\beta$. We therefore remove the $\arctan$ from the equation for $\beta$, as it is computationally expensive and can not easily be provided as a lookup table due to the two float variables $d_1$ and $d_2$. We also provide the sine and cosine of $\alpha$ as constant pre-calculated values. The proper handling of the quadrants of the Euclidean plane usually requires the atan2 function as opposed to the $\arctan$ (and therefore the tangent) \cite{Bogoslavskyi2016}. We circumvent this restriction as the provided depth measurements are always positive or zero, any zero value is automatically removed as ground/invalid value. These pre-calculations change the equation for the horizontal plane estimation to a set of multiplications and each one division and subtraction:
\begin{equation}
\tan(\beta) = \frac{d_2 \cdot \overbrace{\sin{\alpha}}^{\text{Constant}}}{d_1 - d_2 \cdot  \underbrace{\cos{\alpha}}_{\text{Constant}}}.
\label{equation1_simple}
\end{equation}
This method rejects all horizontal surfaces. To prevent excessive removal of valid measurements from elevated horizontal surfaces such as car roofs or hoods, we use a height image, in which the range values are replaced by the Cartesian z-coordinate in relation to the ego vehicle. We use this image to keep all horizontal surfaces above a certain height, in the above case, the line from the wheel position of the ego vehicle to the maximum possible elevation spanned by the 10$^\circ$ slope threshold. A comparable metric has been described by Chu et al. \cite{chu2017fast} although with relative height thresholds, we have decided to use absolute values to reduce the computation time. Figure \ref{groundAngle}, top, depicts the relationship of the channel angles $\delta_r$ (angle in relation to the horizontal $0^\circ$ line) to the surface angles $\beta_r$ (angle of the tangent of two neighbouring measurements and the sensor origin).
\subsection{Clustering}
\label{chapter:methods:clustering}
We use a systematic approach to create an object instance segmentation for Lidar sensor data through clustering in the image space of the range image. This approach relies on the removal of the Lidar measurements belonging to the ground plane from the range image. Inspired by the work of Bogoslavskyi and Stachniss \cite{Bogoslavskyi2016}, we exploit the neighbourhood relationship of adjacent measurements in the range image. As visualised in the bottom illustration of Figure \ref{groundAngle}, we compare the given range values $\vert\vert OA\vert\vert$ and $\vert\vert OB\vert\vert$ for each pair of Lidar measurements. We apply the cosine law to calculate the Euclidean distance $D$ in the three-dimensional space using the two-dimensional range image:
\begin{equation}
\begin{split}
D & = \sqrt{\vert\vert OA\vert\vert^2 + \vert\vert OB\vert\vert^2 - 2 \vert\vert OA\vert\vert \cdot \vert\vert OB\vert\vert \cos{\alpha}}\\
& = \sqrt{d_1^2 + d_2^2 - 2 \cdot d_1 \cdot d_2 \cos{\alpha}}.
\end{split}
\label{equation2}
\end{equation}
The $\alpha$ angle between adjacent Lidar measurements is required for the calculation and is usually provided by the manufacturer of the Lidar sensor for both the horizontal and vertical direction. Using the physical distance between two measured points, we define a threshold value between those, which are close enough together to belong to the same object, or too far apart to be considered neighbours on the same object. The distance of neighbouring points on a given object is in general relatively close. The distances of those points in the range image from two separate objects are substantially larger. By exclusively using variables which are given by the range measurements and reducing the computational effort by pre-calculating the cosine of the given angles, we reduce the calculation of the squared Euclidean distance to a total of four scalar multiplications, an addition and a subtraction:
\begin{equation}
D^2  = d_1 \cdot d_1 + d_2 \cdot d_2 - \underbrace{2 \cdot \cos{\alpha}}_{\text{Constant}} \cdot d_1 \cdot d_2.
\label{equation3}
\end{equation}
These efficient operations are decreasing the runtime on embedded hardware.

With the calculated threshold between each measurement, we are able to connect all Lidar points in the range image into separate clusters and background points. 
With the use of the Euclidean distance as a threshold value, we provide a single parameter implementation with a clear physical meaning, which is adaptable to different sensors.

We reached a good performance with a threshold of 0.8 metres as the limit for the connections between two points. This threshold theoretically enables the clustering of three-dimensional objects with e.g. a Velodyne HDL-64E up to 114.59 metres before the measured points are too far apart on a vertical surface. Horizontally connected components can, in theory, be detected up to a distance of 509.3 metres which is more than four times the reliable range for vehicles of 120 metres, defined by the manufacturer.

\subsubsection{Connected-Component Labelling}\label{chapter:methods:clustering:connected}

\begin{figure}
	\centering
	\includegraphics[width=0.3\textwidth]{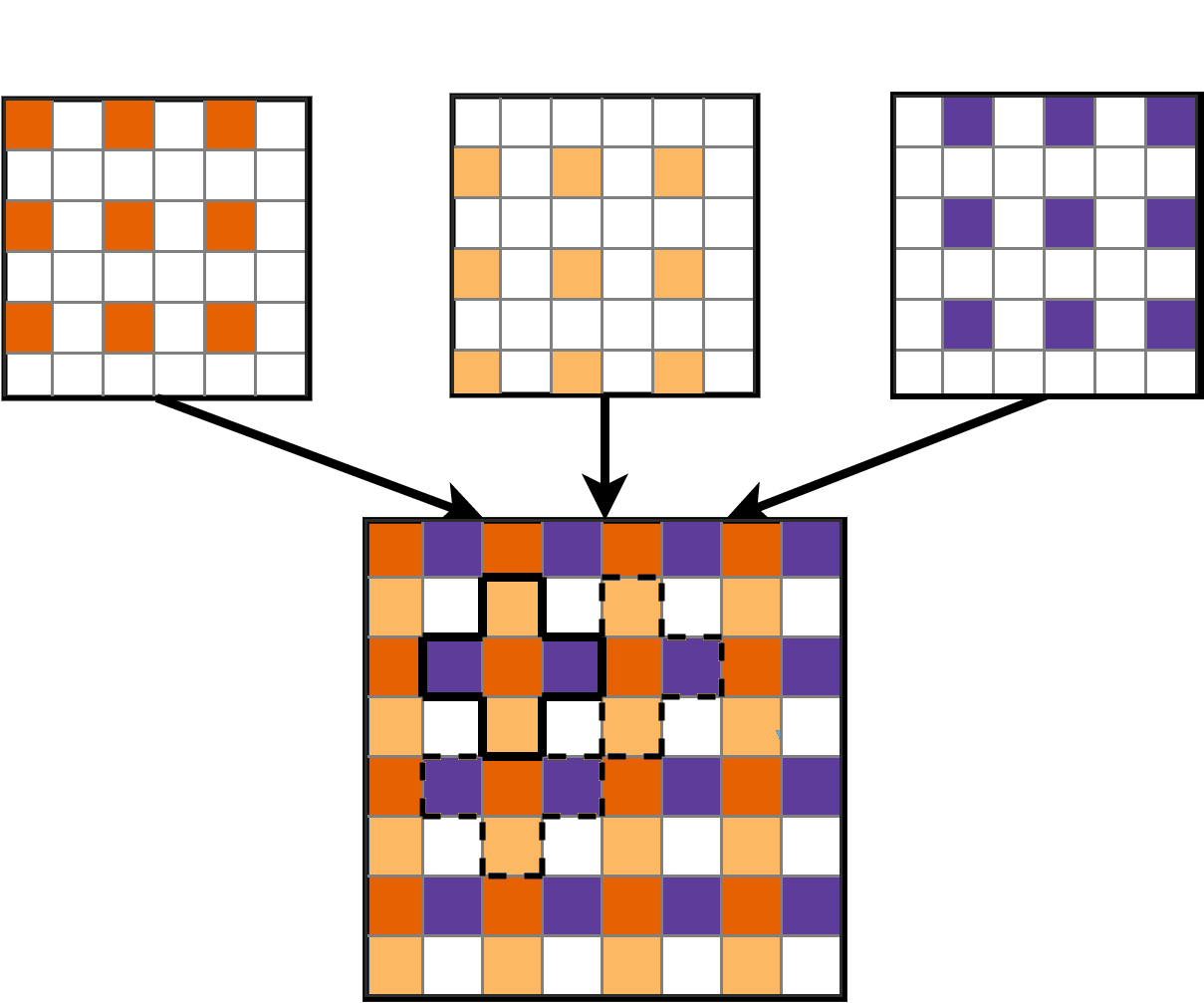}
	\caption{Combination of defined image representations for instance segmentation. The red squares represent the binary value of present Lidar measurements, the yellow and blue squares represent the horizontal and vertical connections of these measurements respectively.}
	\label{supersize}
\end{figure}
Our approach exploits the vectorised nature of the range image to apply operations used in image processing for different purposes. Specifically, we redefine the three-dimensional Euclidean clustering to a two-dimensional connected\--com\-ponent labelling (CCL) problem. To do so, we create two virtual copies of the range image and shift one copy over the $x$ axis and the other one over the $y$ axis. These shifted images enable us to stack all three images on a third axis and compare every value with its vertical and horizontal neighbour over the whole multidimensional array. Thus by using Equation \ref{equation3} on this image we calculate the three-dimensional distance between each point and his vertical and horizontal neighbouring measurement. After applying the threshold on the resulting distance values calculated for each measurement and its direct neighbourhood, we end up with the original range image and two binary images representing the connection or separation between two points in the range image. The range image is now reduced to a binary image representing the presence and absence of Lidar measurements for the corresponding pixels in the image.

The three created binary images contain all the required information to segment the Lidar measurements of the whole frame into clusters and background points. For this, we utilise a simple and efficient image processing algorithm; connected-component labelling. The 4-connected pixel connectivity, also known as von Neumann neighbourhood \cite{toffoli1987cellular}, is defined as a two-dimensional square lattice composed of a central cell and its four adjacent cells. To apply the pixel connectivity to our data, we combine the binary Lidar measurements with the binary threshold images of the distances between Lidar points. By arranging these three images as shown in Figure \ref{supersize}, we are able to apply CCL algorithms with a 4-connectivity on the resulting image to label each island of interconnected measurements as a different cluster. The resulting segmented image is now subsampled to the original range image. Thus we provide the three-dimensional cluster labels directly from the connected-component image, as each pixel corresponds to a given Lidar point in the three-dimensional point cloud.

There is a multitude of CPU-based implementations for CCL problems most common are the "one component at a time" \cite{abubaker2007one} and the Two-pass algorithm \cite{hoshen1976percolation}. We have decided to use the first method in the straight-forward implementation of the \textit{scipy} library "label" for n-dimensional images \cite{virtanen2020scipy}, as it provides a fast \textit{cython} based function. More recent CCL algorithms make use of GPUs by applying the labels in parallel \cite{hennequin2018new} \cite{allegretti2019block}. This can be a very promising approach, as all previous processing steps in this work are applied to rasterised images and can be directly computed in parallel on a GPU. We have not attempted this approach, as our goal is a real-time application for CPU-based automotive hardware.

In a subsequent step we apply a threshold on the labelled clusters for objects below a certain number of Lidar measurements to reduce false clusters resulting from noise in the sensor, in our case we decided on a minimum of 100 points to be considered a cluster candidate, as our objects of interest are cars, pedestrians and other road users. Lower thresholds are recommended to include static objects such as poles and debris on the road.

We have thus segmented the measurements into connected components of separate objects and non-segmented points, which correspond to the ground plane and background noise.

\subsubsection{Map Connections}
\label{chapter:methods:clustering:map}

\begin{figure}[tb]
	\centering
	\vspace{0.4cm}
	\includegraphics[width=0.5\textwidth]{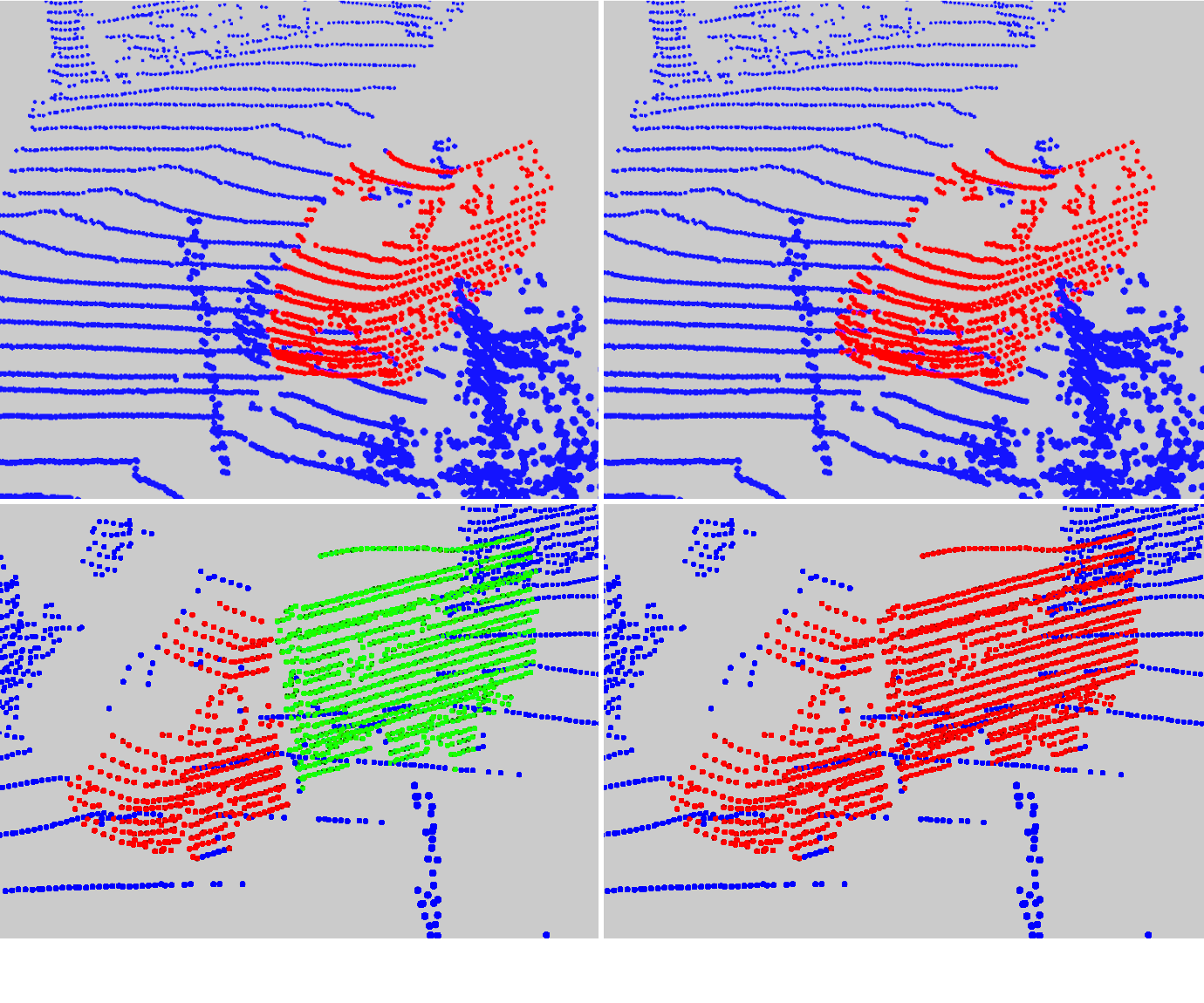}
	\caption{Left: Results using only the direct connectivity between neighbouring Lidar points. Right: A single additional MC between every second Lidar measurement. The proposed MCs enable a more accurate segmentation of the car (top) and reduce the over-segmentation of partially occluded objects, as the truck in the bottom images shows.}
	\label{SkipResults}
\end{figure}

Segmentation algorithms are prone to under- and over-segmentation, due to the characteristics of Lidar sensors; namely the sparsity (especially in vertical direction) and missing measurements resulting from deflected laser beams, which have no remission value back to the sensor. Missing values result in missing connections between areas of the same object, due to which the direct neighbourhood approach described above will over-segment a single object into multiple clusters. Examples of such challenging instances are shown in Figure \ref{SkipResults}.

\begin{figure}[tb]
	\centering
	\includegraphics[width=0.5\textwidth]{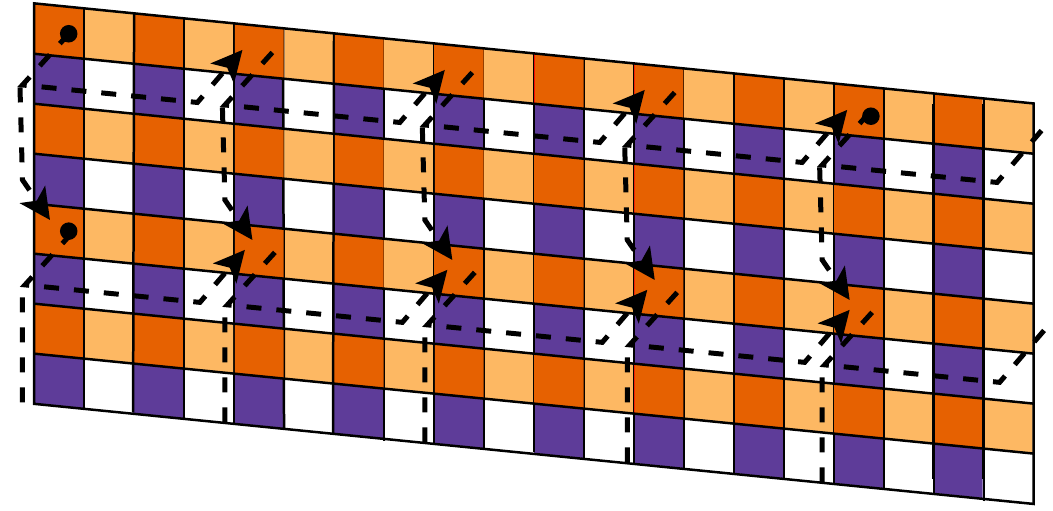}
	\caption{Additional Map Connections (dotted lines) between non-neighbouring Lidar points on top of the direct connections to neighbouring points (yellow and blue squares).}
	\label{SkipConnections}
\end{figure}

To overcome the limitations of the direct neighbourhood approach and to ensure a more robust segmentation, we have extended the two-dimensional Euclidean clustering by what we call \textit{Map Connections} (MC). 
For this, we reduce the combined image shown in Figure \ref{supersize} to a sparse matrix, connecting only every $n^{th}$ point in the vertical and horizontal direction, thus connecting a subset of original points. 
The schematic visualisation in Figure \ref{SkipConnections} displays a connection of each measurement with its second neighbour. Due to the known $\alpha$ angle between all measurements, we can extend the Euclidean distance calculation from each measurement to any other using the cosine law described in Equation \ref{equation3}, by adjusting the angle to the given offset. This allows us to robustly connect segments of the same object, which have no direct connection due to missing measurements or obstruction by other objects in the range image.

\begin{figure}[tb]
	\centering
	\vspace{0.4cm}
	\minipage{0.15\textwidth}
	\includegraphics[width=\textwidth]{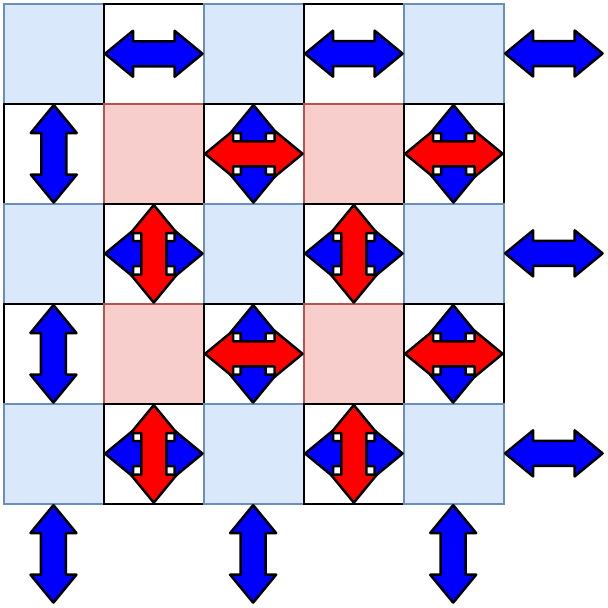}
	\endminipage\hfill
	\minipage{0.15\textwidth}
	\includegraphics[width=\textwidth]{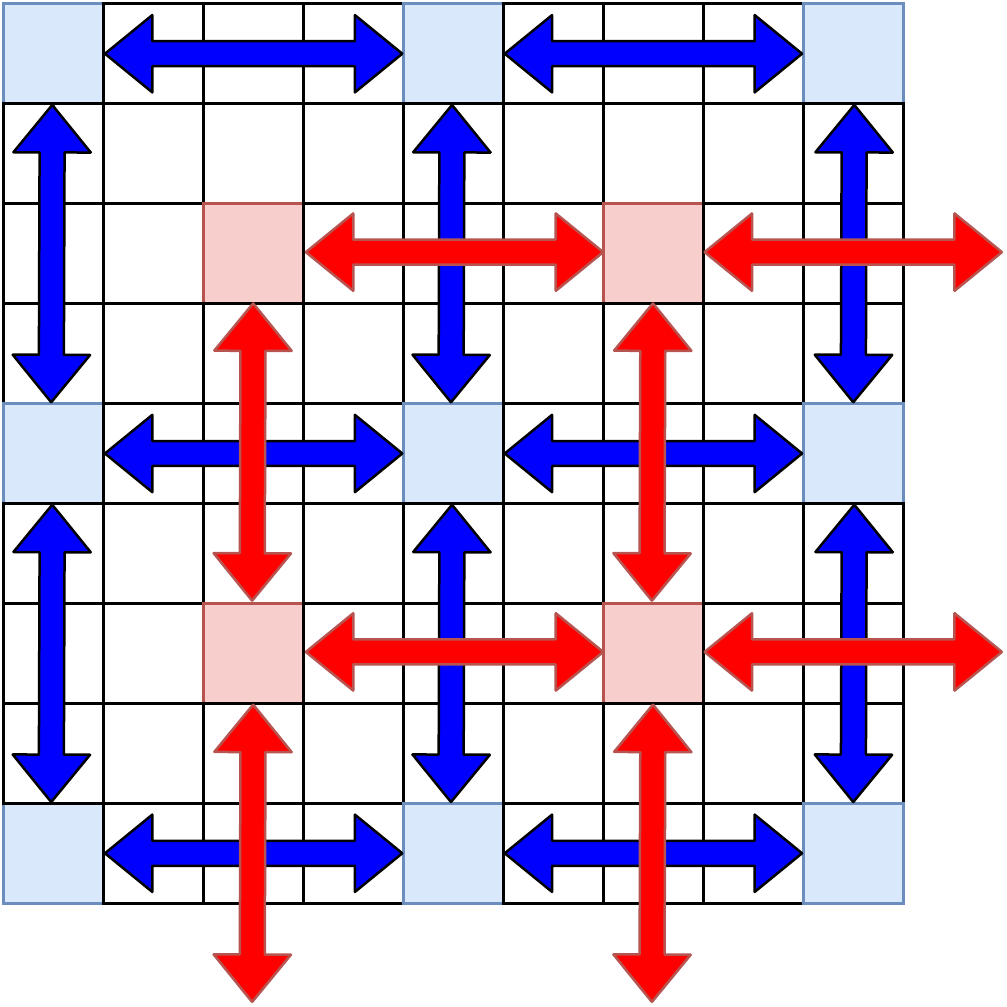}
	\endminipage\hfill
	\minipage{0.15\textwidth}
	\includegraphics[width=\textwidth]{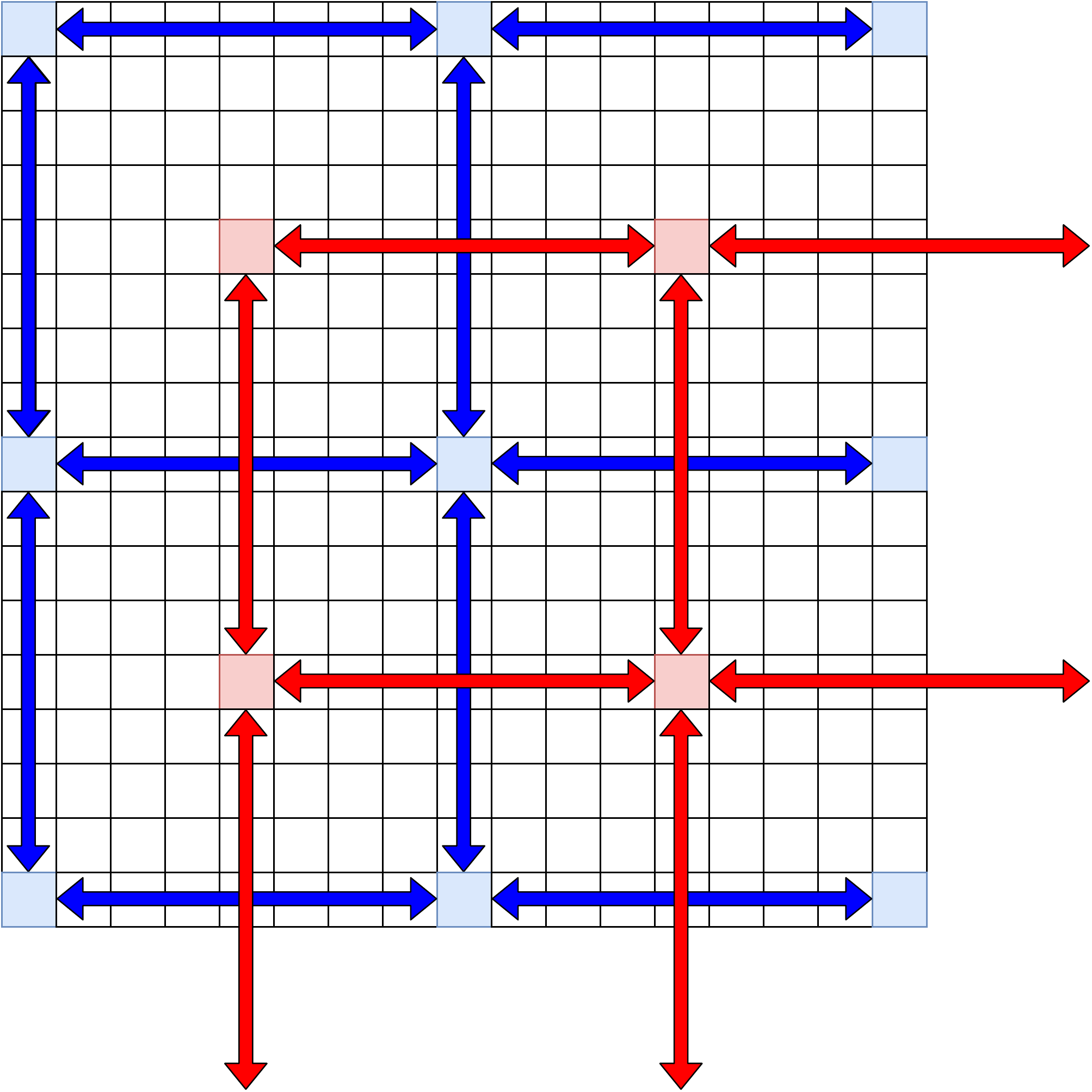}
	\endminipage\hfill
	\vspace{0.3cm}
	\caption{Visualisation of the 6 Map Connection structure for increasing the connection area with the least amount of maps.}
	\label{6maps}
\end{figure}

An example of this improved segmentation can be seen in Figure \ref{SkipResults}. In the results shown in Section \ref{chapter:evaluation} we have added one MC between every second measurement as sketched in Figure \ref{SkipConnections}, 6 MCs as shown in Figure \ref{6maps} and 14 MCs along the main diagonal of the range image. The \textit{Maps} of reduced point-sets are smaller than the original point-set and thus require only a fraction of the computation time on top of the directly connected clusters. The additional mapping of the cluster-ids of the original clusters with the MCs, results in a slightly increased runtime as shown in Section \ref{chapter:evaluation:runtime}.
The combined use of the direct connectivity of neighbouring measurements and the MCs enables a pseudo three-dimensional Euclidean clustering while exploiting the fast runtime of two-dimensional pixel connectivity. Thus, we are able to improve the quality of the segmentation without sacrificing our real-time ability. 

\section{Experimental Evaluation}\label{chapter:evaluation}
\noindent The first experimental evaluation measures our methods ability to run in real time at usual sensor recording frequencies, while offering a constant processing rate with very little fluctuation independent of the scene's context. The second experiment is mainly concerned with a quantitative metric of the segmentation quality.

\subsection{Runtime}
\label{chapter:evaluation:runtime}

Following the experimental setup of \cite{Bogoslavskyi2016}, we designed our first experiment on the provided data by Moosmann et al. \cite{moosmann2013interlacing} to support the claim, that the proposed approach can be used for online segmentation. All listed methods have been evaluated on the same Intel\textsuperscript{\textregistered} Core\texttrademark \: i7-6820HQ CPU @ 2.70 GHz.

\begin{figure*}[tb]
	\centering
	\includegraphics[width=\textwidth]{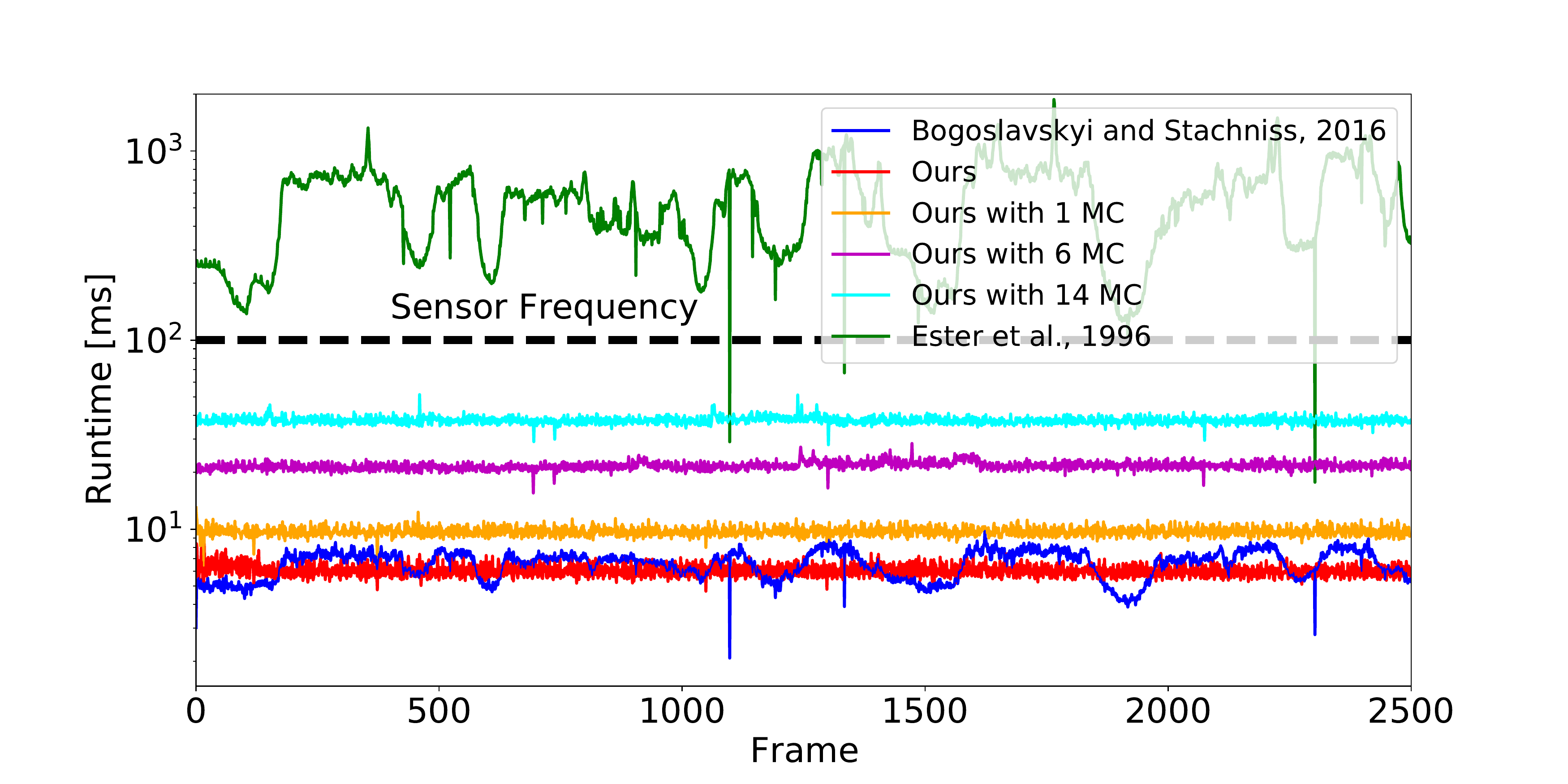}
	\caption{Frame-wise execution timings on a 64-beam Velodyne dataset \cite{moosmann2013interlacing}. Please note the logarithmic scale for the runtime.}
	\label{fig:timeplot}
\end{figure*}

\begin{figure*}[tb]
	\centering
	\includegraphics[width=\textwidth]{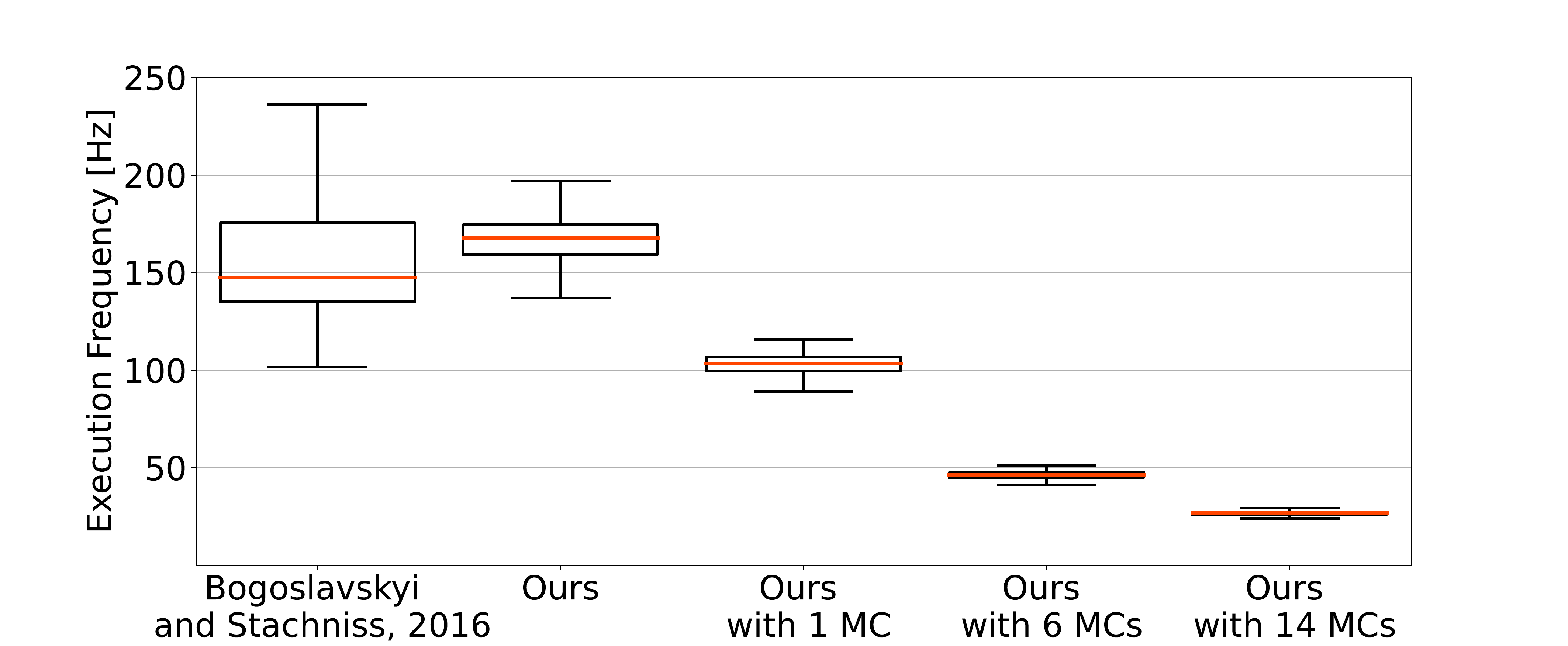}
	\caption{Averaged runtime in Hz for segmenting approximately 2,500 scans from a 64-beam Velodyne dataset \cite{moosmann2013interlacing} with our approach and up to 14 Map Connections compared to the method by Bogoslavskyi and Stachniss \cite{Bogoslavskyi2016}.}
	\label{fig:boxplot}
	
\end{figure*}

Figure \ref{fig:timeplot} shows the execution time of the 5 methods over the 2500 Frames dataset \cite{moosmann2013interlacing}. The proposed method runs at an average of 165 Hz and is therefore faster than the previously fastest algorithm of \cite{Bogoslavskyi2016} at 152 Hz, while exhibiting less fluctuation due to the binary image implementation when used without any additional MCs. A box-plot of the average runtime of \cite{Bogoslavskyi2016} and the proposed method can be seen in Figure \ref{fig:boxplot}, which shows the fluctuating nature of methods depending on the scene context, as opposed to ours.
\begin{table*}[tb]
	\vspace{2.5mm}
	\begin{center}
		\begin{tabular}[width=\textwidth]{lrrr|lrrrr}
			\setlength{\tabcolsep}{10pt}
			
			\textbf{Method}  & $IoU_{\mu}$  & $IoU_{\mu}$ &&& $P_{\mu}$ & $P_{0.5}$ & $P_{0.75}$ & $P_{0.95}$\\ 
			& (No Ground) & (Ground) & & & \hspace{1cm} & \hspace{1cm} & \hspace{1cm} & \hspace{1.5cm}\\
			\hline
			\hline
			&&&&&&&&\\
			Bogoslavskyi et al. & 73.93          & 73.93          &&& 59.31          & 83.75          & 63.52          & 13.18\\
			Ours                                        & 76.20          & 72.31          &&& 63.73          & 84.30          & 67.51          & 22.03 \\ 
			Ours (\textit{1 MC})                        & 77.97          & 73.65          &&& 66.68          & 85.60          & 70.21          & 27.19 \\		
			Ours (\textit{6 MC})                        & 81.14 & 75.48          &&& 71.92          & 88.25          & 74.99          & 36.05 \\ 
			Ours (\textit{14 MC})                       & \textbf{84.25} & \textbf{76.39} &&& 74.68 & \textbf{89.75} & \textbf{77.61} & 40.63 \\ 
			DBSCAN       & 76.21          & 72.77          &&& \textbf{76.50}          & 81.54          & 76.45          & \textbf{69.25} \\
			\hline
		\end{tabular} 
	\end{center}
	\caption{Comparison of the segmentation quality using the Intersection over Union and the precision average for the algorithms of \cite{Bogoslavskyi2016}, \cite{ester1996density}, and variations of the proposed method}
\label{table:iouP}
\end{table*}
As can be seen in Figure \ref{fig:timeplot}, when adding MCs to the proposed methods, the execution suffers from a slightly longer runtime, while still running at a frequency of 26 to 105 Hz depending on the number of additional MCs. This is still between 2.6 to ten times faster than the recording frequency of the used sensor. Please note that we only used up to 14 MC in this timing. More MCs increase the execution time accordingly and endanger the real time capability of the proposed method. Implementations of the \textit{Map Connection} assignment in an efficient programming language like \textit{C++} might enable the use of more MCs with less influence on the execution time.

\subsection{Segmentation Results}
\label{chapter:evaluation:quality}

For our evaluation, we use the dataset ``SemanticKITTI" by Behley et al. \cite{behley2019semantickitti} as object instance segmentation ground truth. This dataset enriches the KITTI dataset's \cite{Geiger2012} odometry challenge with semantic and instance-wise labels for every Lidar measurement.

To reduce the influence of the proposed ground plane extraction in Section \ref{chapter:methods:ground} and focus on the results of the clustering mechanisms, we have conducted the evaluation once without the Lidar points of the classes road, parking, sidewalk, other-ground, lane-marking and terrain. And a second time without any usage of the semantic labels by applying the ground extraction proposed in \ref{chapter:methods:ground} on all methods.

For each ground truth object with at least 100 Lidar point measurements, we select each algorithm's object cluster output with the most ground truth overlap. Using these two lists of points, we calculate the Intersection over Union (IoU). By averaging these IoU values of every single instance over all 10 sequences, we get the averaged mean $IoU_{\mu}$ of each algorithm. The IoU or \textit{Jaccard Index} for a single ground-truth instance with a single cluster is defined as
\begin{equation}
J(A, B) = \frac{\vert A \cap B \vert}{\vert A \cup B \vert} \:\: \widehat{=} \:\: \frac{TP}{TP+FP+FN}.
\end{equation}
The connection and separation of instances solely through the distance harbours the risk of under-segmentation in the case of objects that are in contact or are close by. For this reason, we measure our results in the subsequent evaluation instance-wise. If two instances are represented by only one cluster, we count only the object with the higher IoU, while the second object is marked as not found. We compute this metric for each algorithm listed below, over all ten sequences with Lidar instance ground truth in the dataset. We compare the quality of our algorithm to the currently fastest algorithm \cite{Bogoslavskyi2016} as well as a very precise three-dimensional euclidean clustering algorithm \cite{ester1996density}. We use scikit-learn’s implementation of DBSCAN \cite{scikit-learn}. If you consider the age of this algorithm, it might be surprising to see that it is still used in modern clustering applications \cite{zhangrobust} \cite{mao2020pedestrian} \cite{cheng2019pedestrian}. This long-term relevance was also confirmed by the "Test of Time" Award from ACM SIGKDD \cite{testoftime}. The algorithm was also revisited by the original authors in a follow-up paper \cite{schubert2017dbscan} fairly recently to show the continued relevance in many clustering applications. Therefore we use the DBSCAN to compare our pseudo three-dimensional approach to a well performing three-dimensional algorithm.

\begin{figure}[tb]
	\centering
	\includegraphics[width=0.52\textwidth]{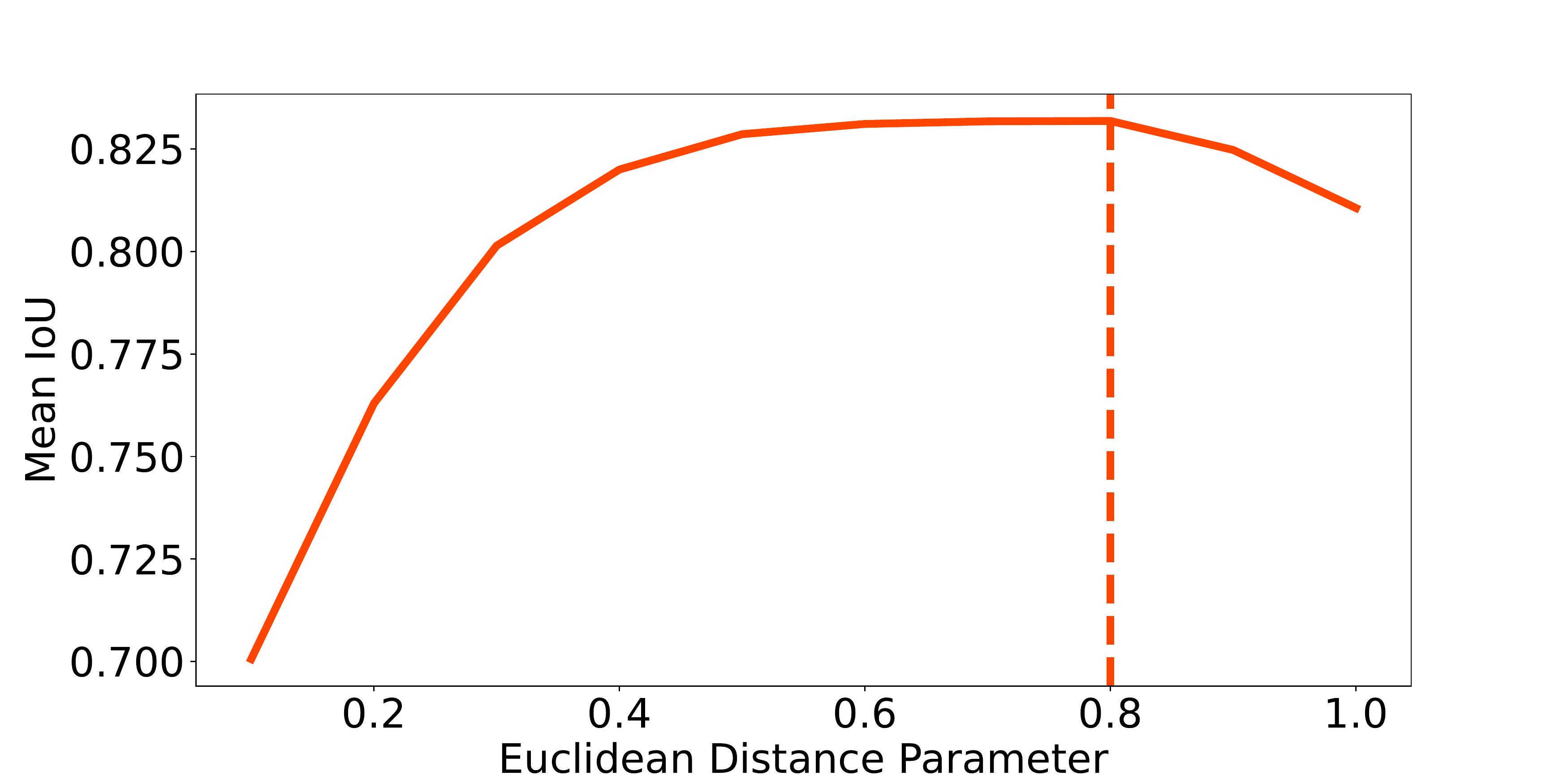}
	\caption{Parameter study of the maximum distance between two points, to be considered part of the same cluster. The dashed line shows the maximum IoU for the evaluation log. The plateau between $\unit{0.5m}$ and $\unit{0.8m}$ shows a very broad and robust sweet spot for the proposed method.}
	\label{fig:euclidean_parameter}
\end{figure}

We present the mean of the IoUs with the best-performing parameters of each method. With our threshold parameter set as $\unit{0.8m}$, we outperform \cite{Bogoslavskyi2016} with the direct neighbourhood implementation without any MCs while exhibiting a faster run-time. This parameter has been set with an additional experiment on a single log of the dataset as can be seen in Figure \ref{fig:euclidean_parameter}. Other Lidar sensors perform better with a higher or lower threshold depending on the horizontal and vertical resolution of the Lidar sensor. We recommend to evaluate the euclidean distance parameter for the use of different sensors.
On average our method is 8\% faster, while the difference in the respectively longest execution time of a given frame is 15\%. Together with the proposed MCs between all odd measurements, we manage to perform higher on the average IoU than the more general three-dimensional Euclidean distance clustering algorithm DBSCAN \cite{ester1996density}, as shown in Table \ref{table:iouP}. 

Using 6 MCs, we surpass the performance of DBSCAN with a larger margin and manage to reach a noticeably higher mean IoU. A total of 14 MCs outperforms the three-dimensional DBSCAN algorithm on 4 of the 6 shown metrics in Table \ref{table:iouP}, with an average execution frequency of 26 Hz it still runs at 2.6 times the sensor frequency. We suspect that the restriction to the immediate surroundings of the measurements prevents under-segmentation of separate, close objects, but still enables the skipping of gaps on the same object. 

Without any MCs we are on average 120 times faster. With an increasing number of additional MCs we are 67, 25 and 14 times faster than the DBSCAN Algorithm. Even if we drastically increase the number of MCs, our proposed method is at least 14 times faster than the DBSCAN and remains clearly above the sensor recording frequency.The run-time increase does not scale logarithmically as one would expect with additional MCs (since they apply the same function to a smaller subset of the original point cloud). This issue might result from the mapping overhead caused by our python implementation. We did not re-implement the method in a different programming language, as the computation time was still far below the sensor frequency and does not bring any further benefits to do so.

The second column of Table \ref{table:iouP} shows, that the proposed ground extraction method of Section \ref{chapter:methods:ground} degrades the performance of all listed algorithms, except for \cite{Bogoslavskyi2016} as the used metric for the ground segmentation is very similar to the cluster separation metric used by Bogoslavskyi et al. A better ground separation will lead to a much better performance for the proposed method as the IoU values with the ground truth (GT) ground segmentation show. Improving the ground separation is therefore critical to improve the instance segmentation.

For further evaluation of the instance-level performance, we compute the precision similar to \cite{Bogoslavskyi2016} in a more refined fashion on the GT ground removed point clouds.
We define 10 bins of point-wise overlap of the GT and proposed clusters ranging from an IoU of 0.5 to 0.95 in steps of 0.05. We average the precision of all bins into one single metric score ($P_{\mu}$) which is shown in Table \ref{table:iouP} for each method. We additionally list the Precision for the overlap values of 0.5, 0.75 and 0.95, in which the definition of a correctly segmented object is defined as above the denoted IoU.
The precision, which shows how many instances are matched with an IoU of at least x is therefore defined as

\begin{equation}
\begin{aligned}
&P_x = \frac{1}{N} \sum_{n=0}^{N}\sum_{m=0}^{M}  a_{n,m} \\ 
&\textnormal{with } a_{n,m}  = \begin{cases}
1  , \textnormal{ if } J(n,m) >= x, \\  
0  , \textnormal{ else.} \\  
\end{cases} \\
\end{aligned}
\end{equation}

for N instances and M clusters in which each instance and cluster are matched via the \textit{Jaccard Index} (IoU). Please note, that due to the definition of the \textit{Jaccard Index} only one cluster can match a ground truth instance with an $IoU > 0.5$.

We show in Table \ref{table:iouP}, that we match on average more GT instances than \cite{Bogoslavskyi2016} and are close to the mean segmentation precision of the DBSCAN algorithm. With a higher number of MCs, we achieve better precision values for overlap values of 0.5 and 0.75, while the DBSCAN algorithm matches more instances with higher overlap values due to the full three-dimensional clustering on all points of the dataset. 

We only compare up to 14 MCs in order to not endanger our real-time capability. However, with just these 14 MCs, we achieve a clustering segmentation which performs comparable to, and in some regards better, than the full three-dimensional algorithm.
The high precision values for the lower overlap regions of 0.5 and 0.75 are particularly important in the context of driver assistance systems, since a missed instance can lead to dramatic outcomes, as opposed to a not perfectly matched instance. We further proved, that the proposed MCs improve the results of our algorithm immensely and help to find otherwise missed objects. 

Attentive readers will notice the performance of both our method and the method by \cite{Bogoslavskyi2016} drop noticeably in the 0.95 IoU bracket. We see this issue to be due to the underlying data. The ``SemanticKITTI" dataset \cite{behley2019semantickitti} has a pre-applied ego-motion compensation, due to which the three-dimensional point cloud is slightly shifted and rotated away from the original sensor configuration to compensate the movement in the $\unit{0.1s}$ recording time of a single frame. This compensation has already been applied to the original odometry dataset of Geiger et al. \cite{Geiger2012} to provide better Lidar odometry for static inter-frame point cloud matching.

Our method builds on this three-dimensional point cloud and expects an unaltered version of the Lidar sensor to enable a one-to-one matching of the three dimensional points to the two-dimensional range image representation. This can not be fully achieved with this ego-motion compensated data and results in missing and wrongly assigned points and in turn hurts our performance in the 0.95 IoU bracket, which requires a precise projection. We see the same drop amplified in the range-image based algorithm of \cite{Bogoslavskyi2016}, while the DBSCAN algorithm runs directly on the manipulated three-dimensional data and does not suffer from this restriction.

\section{Conclusion}

\noindent We have presented an algorithm for real-time instance segmentation of Lidar sensor data using raw range images to connect points by their three-dimensional distance. To make this approach more robust against over-segmentation, we introduced what we call \textit{Map Connections}, which use the larger neighbouring context for a more precise assignment of measured points to an instance, especially in cases of partial occlusion. These properties of our method facilitate the preservation of three-dimensional information in the measurements when reduced to a two-dimensional representation for fast computation.

In a detailed evaluation, we have shown, that our approach is faster than comparable state-of-the-art methods, while being more stable in its runtime, and more importantly, providing an overall better performance in instance segmentation. The experiments show, that not only our accuracy in separating objects is higher than comparable fast approaches, but we are able to match most ground truth instances with a significant overlap of the ground truth. This is particularly important in the context of driver assistance systems, since a missed instance is a bigger problem than an object that was not matched with all Lidar points. We have also illustrated, that the proposed MCs improve the results of our algorithm and help to find otherwise missed objects. The further use of segmented point clouds for classification and to remove false positives, is outside the scope of this work. However, this application has previously been researched by \cite{hahn2020fast} and shows promising results.






\newpage
\bibliography{Bibliography}

\begin{thebibliography}{10}

\bibitem{ester1996density}
M.~Ester, H.-P. Kriegel, J.~Sander, X.~Xu, {\em et~al.}, ``A density-based
  algorithm for discovering clusters in large spatial databases with noise.,''
  in {\em Kdd}, vol.~96, pp.~226--231, 1996.

\bibitem{fukunaga1975estimation}
K.~Fukunaga and L.~Hostetler, ``The estimation of the gradient of a density
  function, with applications in pattern recognition,'' {\em IEEE Transactions
  on information theory}, vol.~21, no.~1, pp.~32--40, 1975.

\bibitem{comaniciu2002mean}
D.~Comaniciu and P.~Meer, ``Mean shift: A robust approach toward feature space
  analysis,'' {\em IEEE Transactions on pattern analysis and machine
  intelligence}, vol.~24, no.~5, pp.~603--619, 2002.

\bibitem{ankerst1999optics}
M.~Ankerst, M.~M. Breunig, H.-P. Kriegel, and J.~Sander, ``Optics: ordering
  points to identify the clustering structure,'' {\em ACM Sigmod record},
  vol.~28, no.~2, pp.~49--60, 1999.

\bibitem{himmelsbach2010fast}
M.~Himmelsbach, F.~V. Hundelshausen, and H.-J. Wuensche, ``Fast segmentation of
  3d point clouds for ground vehicles,'' in {\em 2010 IEEE Intelligent Vehicles
  Symposium}, pp.~560--565, IEEE, 2010.

\bibitem{korchev2013real}
D.~Korchev, S.~Cheng, Y.~Owechko, {\em et~al.}, ``On real-time lidar data
  segmentation and classification,'' in {\em Proceedings of the International
  Conference on Image Processing, Computer Vision, and Pattern Recognition
  (IPCV)}, p.~1, The Steering Committee of The World Congress in Computer
  Science, Computer~…, 2013.

\bibitem{moosmann2009segmentation}
F.~Moosmann, O.~Pink, and C.~Stiller, ``Segmentation of 3d lidar data in
  non-flat urban environments using a local convexity criterion,'' in {\em 2009
  IEEE Intelligent Vehicles Symposium}, pp.~215--220, IEEE, 2009.

\bibitem{Bogoslavskyi2016}
I.~Bogoslavskyi and C.~Stachniss, ``Fast range image-based segmentation of
  sparse 3d laser scans for online operation,'' in {\em 2016 IEEE/RSJ
  International Conference on Intelligent Robots and Systems (IROS)},
  pp.~163--169, IEEE, 2016.

\bibitem{zermas2017fast}
D.~Zermas, I.~Izzat, and N.~Papanikolopoulos, ``Fast segmentation of 3d point
  clouds: A paradigm on lidar data for autonomous vehicle applications,'' in
  {\em 2017 IEEE International Conference on Robotics and Automation (ICRA)},
  pp.~5067--5073, IEEE, 2017.

\bibitem{lahoud20193d}
J.~Lahoud, B.~Ghanem, M.~Pollefeys, and M.~R. Oswald, ``3d instance
  segmentation via multi-task metric learning,'' in {\em Proceedings of the
  IEEE International Conference on Computer Vision}, pp.~9256--9266, 2019.

\bibitem{zhang2020instance}
F.~Zhang, C.~Guan, J.~Fang, S.~Bai, R.~Yang, P.~Torr, and V.~Prisacariu,
  ``Instance segmentation of lidar point clouds,'' {\em ICRA}, vol.~4, no.~1,
  2020.

\bibitem{yang2019learning}
B.~Yang, J.~Wang, R.~Clark, Q.~Hu, S.~Wang, A.~Markham, and N.~Trigoni,
  ``Learning object bounding boxes for 3d instance segmentation on point
  clouds,'' in {\em Advances in Neural Information Processing Systems},
  pp.~6740--6749, 2019.

\bibitem{wang2019ldls}
B.~H. Wang, W.-L. Chao, Y.~Wang, B.~Hariharan, K.~Q. Weinberger, and
  M.~Campbell, ``Ldls: 3-d object segmentation through label diffusion from 2-d
  images,'' {\em IEEE Robotics and Automation Letters}, vol.~4, no.~3,
  pp.~2902--2909, 2019.

\bibitem{wang2019pointit}
Y.~Wang, Y.~Yu, and M.~Liu, ``Pointit: A fast tracking framework based on 3d
  instance segmentation,'' {\em arXiv preprint arXiv:1902.06379}, 2019.

\bibitem{behley2019semantickitti}
J.~Behley, M.~Garbade, A.~Milioto, J.~Quenzel, S.~Behnke, C.~Stachniss, and
  J.~Gall, ``Semantickitti: A dataset for semantic scene understanding of lidar
  sequences,'' in {\em Proceedings of the IEEE International Conference on
  Computer Vision}, pp.~9297--9307, 2019.

\bibitem{kochanov2020kprnet}
D.~Kochanov, F.~K. Nejadasl, and O.~Booij, ``Kprnet: Improving projection-based
  lidar semantic segmentation,'' {\em arXiv preprint arXiv:2007.12668}, 2020.

\bibitem{zhou2020cylinder3d}
H.~Zhou, X.~Zhu, X.~Song, Y.~Ma, Z.~Wang, H.~Li, and D.~Lin, ``Cylinder3d: An
  effective 3d framework for driving-scene lidar semantic segmentation,'' {\em
  arXiv preprint arXiv:2008.01550}, 2020.

\bibitem{zhang12356deep}
F.~Zhang, J.~Fang, B.~Wah, and P.~Torr, ``Deep fusionnet for point cloud
  semantic segmentation,''

\bibitem{tang2020searching}
H.~Tang, Z.~Liu, S.~Zhao, Y.~Lin, J.~Lin, H.~Wang, and S.~Han, ``Searching
  efficient 3d architectures with sparse point-voxel convolution,'' {\em arXiv
  preprint arXiv:2007.16100}, 2020.

\bibitem{gerdzhev2020tornado}
M.~Gerdzhev, R.~Razani, E.~Taghavi, and B.~Liu, ``Tornado-net: multiview total
  variation semantic segmentation with diamond inception module,'' {\em arXiv
  preprint arXiv:2008.10544}, 2020.

\bibitem{cortinhal2020salsanext}
T.~Cortinhal, G.~Tzelepis, and E.~E. Aksoy, ``Salsanext: Fast semantic
  segmentation of lidar point clouds for autonomous driving,'' {\em arXiv
  preprint arXiv:2003.03653}, 2020.

\bibitem{milioto2020iros}
A.~Milioto, J.~Behley, C.~McCool, and C.~Stachniss, ``{LiDAR Panoptic
  Segmentation for Autonomous Driving},'' 2020.

\bibitem{chu2017fast}
P.~Chu, S.~Cho, S.~Sim, K.~Kwak, and K.~Cho, ``A fast ground segmentation
  method for 3d point cloud.,'' {\em Journal of information processing
  systems}, vol.~13, no.~3, 2017.

\bibitem{toffoli1987cellular}
T.~Toffoli and N.~Margolus, {\em Cellular automata machines: a new environment
  for modeling}.
\newblock MIT press, 1987.

\bibitem{abubaker2007one}
A.~AbuBaker, R.~Qahwaji, S.~Ipson, and M.~Saleh, ``One scan connected component
  labeling technique,'' in {\em 2007 IEEE International Conference on Signal
  Processing and Communications}, pp.~1283--1286, IEEE, 2007.

\bibitem{hoshen1976percolation}
J.~Hoshen and R.~Kopelman, ``Percolation and cluster distribution. i. cluster
  multiple labeling technique and critical concentration algorithm,'' {\em
  Physical Review B}, vol.~14, no.~8, p.~3438, 1976.

\bibitem{virtanen2020scipy}
P.~Virtanen, R.~Gommers, T.~E. Oliphant, M.~Haberland, T.~Reddy, D.~Cournapeau,
  E.~Burovski, P.~Peterson, W.~Weckesser, J.~Bright, {\em et~al.}, ``Scipy 1.0:
  fundamental algorithms for scientific computing in python,'' {\em Nature
  methods}, vol.~17, no.~3, pp.~261--272, 2020.

\bibitem{hennequin2018new}
A.~Hennequin, L.~Lacassagne, L.~Cabaret, and Q.~Meunier, ``A new direct
  connected component labeling and analysis algorithms for gpus,'' in {\em 2018
  Conference on Design and Architectures for Signal and Image Processing
  (DASIP)}, pp.~76--81, IEEE, 2018.

\bibitem{allegretti2019block}
S.~Allegretti, F.~Bolelli, M.~Cancilla, and C.~Grana, ``A block-based
  union-find algorithm to label connected components on gpus,'' in {\em
  International Conference on Image Analysis and Processing}, pp.~271--281,
  Springer, 2019.

\bibitem{moosmann2013interlacing}
F.~Moosmann, {\em Interlacing self-localization, moving object tracking and
  mapping for 3d range sensors}, vol.~24.
\newblock KIT Scientific Publishing, 2013.

\bibitem{Geiger2012}
A.~Geiger, P.~Lenz, and R.~Urtasun, ``Are we ready for autonomous driving? the
  kitti vision benchmark suite,'' in {\em 2012 IEEE Conference on Computer
  Vision and Pattern Recognition}, pp.~3354--3361, IEEE, 2012.

\bibitem{scikit-learn}
F.~Pedregosa, G.~Varoquaux, A.~Gramfort, V.~Michel, B.~Thirion, O.~Grisel,
  M.~Blondel, P.~Prettenhofer, R.~Weiss, V.~Dubourg, J.~Vanderplas, A.~Passos,
  D.~Cournapeau, M.~Brucher, M.~Perrot, and E.~Duchesnay, ``Scikit-learn:
  Machine learning in {P}ython,'' {\em Journal of Machine Learning Research},
  vol.~12, pp.~2825--2830, 2011.

\bibitem{zhangrobust}
S.~Zhang, F.~M. Di~Wang, C.~Qin, Z.~Chen, and M.~Liu, ``Robust pedestrian
  tracking in crowd scenarios using an adaptive gmm-based framework,''

\bibitem{mao2020pedestrian}
J.~Mao, G.~Xu, W.~Li, X.~Fan, and J.~Luo, ``Pedestrian detection and
  recognition using lidar for autonomous driving,'' in {\em 2019 International
  Conference on Optical Instruments and Technology: Optical Sensors and
  Applications}, vol.~11436, p.~114360R, International Society for Optics and
  Photonics, 2020.

\bibitem{cheng2019pedestrian}
H.~Cheng, Y.~Li, and M.~Sester, ``Pedestrian group detection in shared space,''
  in {\em 2019 IEEE Intelligent Vehicles Symposium (IV)}, pp.~1707--1714, IEEE,
  2019.

\bibitem{testoftime}
A.~SIGKDD, ``2014 sigkdd test of time award,'' {\em
  https://www.kdd.org/News/view/2014-sigkdd-test-of-time-award}, 2014.

\bibitem{schubert2017dbscan}
E.~Schubert, J.~Sander, M.~Ester, H.~P. Kriegel, and X.~Xu, ``Dbscan revisited,
  revisited: why and how you should (still) use dbscan,'' {\em ACM Transactions
  on Database Systems (TODS)}, vol.~42, no.~3, pp.~1--21, 2017.

\bibitem{hahn2020fast}
L.~Hahn, F.~Hasecke, and A.~Kummert, ``Fast object classification and
  meaningful data representation of segmented lidar instances,'' {\em 23rd IEEE
  International Conference on Intelligent Transportation Systems (ITSC)}, 2020.

\end{thebibliography}
\bibliographystyle{ieeetr}

\end{document}